\documentclass[conference]{IEEEtran}
\IEEEoverridecommandlockouts
\usepackage{cite}
\usepackage{amsmath,amssymb,amsfonts}
\usepackage{algorithmic}
\usepackage{graphicx}
\usepackage{textcomp}
\usepackage[compact]{titlesec}        
\titlespacing{\section}{1pt}{1pt}{1pt} 
\AtBeginDocument{                     
  \setlength\abovedisplayskip{1pt}
  \setlength\belowdisplayskip{1pt}}
\usepackage{xcolor}
\usepackage{subfig}
\usepackage[ruled,vlined]{algorithm2e}
\def\BibTeX{{\rm B\kern-.05em{\sc i\kern-.025em b}\kern-.08em
    T\kern-.1667em\lower.7ex\hbox{E}\kern-.125emX}}
\begin{document}

\title{A L\'evy Flight based Narrow Passage Sampling Method for Probabilistic Roadmap Planners
}

\author{Shubham Shukla$^{1}$, Lokesh Kumar$^{1}$, Titas Bera $^{1}$, Ranjan Dasgupta$^{1}$ % <-this % stops a space
% \thanks{${\dagger}$ Represents equal contribution}% <-this % stops a space
\thanks{$^{1}$TCS Research \& Innovations Lab, Kolkata, India.\newline
         {\tt\small \{shukla.shubh, kumar.lokesh7, titas.bera, ranjan.dasgupta\}@tcs.com}}%
}
\maketitle

\begin{abstract}
Sampling based probabilistic roadmap planners (PRM) have been successful in motion planning of robots with higher degrees of freedom, but may fail to capture the connectivity of the configuration space in scenarios with a critical narrow passage. In this paper, we show a novel technique based on L\'evy Flights to generate key samples in the narrow regions of configuration space, which, when combined with a PRM, improves the completeness of the planner. The technique substantially improves sample quality at the expense of a minimal additional computation, when compared with pure random walk based methods, however, still outperforms state of the art random bridge building method, in terms of number of collision calls, computational overhead and sample quality. The method is robust to the changes in the parameters related to the structure of the narrow passage, thus giving an additional generality. A number of $2$D \& $3$D motion planning simulations are presented which shows the effectiveness of the method. 
\end{abstract}

\section{Introduction}
It is known that the basic problem of robot path planning is PSPACE-complete \cite{schwartz1983piano} and exact deterministic algorithms exhibits curse of dimensionality problem \cite{reif1979complexity}. To resolve such, several sampling based probabilistic motion planners have been proposed and widely used successfully for path planning for robots in high dimensional configuration spaces such as  Probabilistic Roadmaps (PRM) \cite{kavraki1996probabilistic}, Rapidly-Exploring Random Trees (RRT)\cite{lavalle2001rapidly}, etc.

Although the success of sampling based path planners are documented for various complicated scenario, path planning in high dimensional configuration space is still a difficult task. For sampling based planners one such difficulty arises when the corresponding configuration space possesses narrow passages. As the probability of sampling candidate configurations within narrow regions are proportional to the Lebesgue measure of the narrow passage region,  this makes it difficult to obtain sample points in narrow passages, where the measure of such region is very less. The failure of generating candidate configuration within the narrow region may adversely affect the connectivity of the random graph structure. Several heuristic based sampling strategies can remove the difficulty to a great extent, but a satisfactory answer remains elusive.

In this paper, we propose a novel L\'evy Flight (LFS) based bridge sampling method, a modification over randomized bridge builder mechanism, to address the narrow passage sampling problem. Using a L\'evy Flight mechanism \cite{shlesinger1995levy}, the method is able to generate an improved quality sample configurations near the critical regions with a reduced computational cost when compared to  state-of-the-art algorithms. We compare these with the standard Random walk to surface \cite{bera2010efficient}, Gaussian Sampler \cite{boor1999gaussian}, and the vanilla Randomized Bridge Builder (RBB) mechanism \cite{hsu2003bridge} to show superiority of our proposed algorithm in terms of improved sample quality and reduced time taken. For completeness, we integrated the proposed sampling scheme into a PRM like planner and show the motion trajectories for various scenarios.

\begin{figure} 
    \centering
    \subfloat[]{\includegraphics[width=4cm, height=3cm]{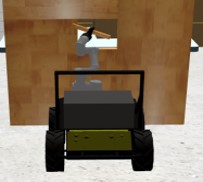}}\hspace{0.25cm}
    \subfloat[]{\includegraphics[width=4cm, height=3cm]{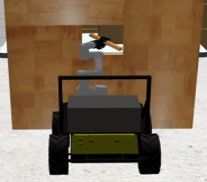}}\\  
    \subfloat[]{\includegraphics[width=4cm, height=3cm]{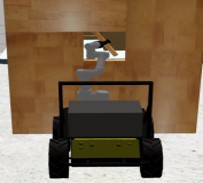}}\hspace{0.25cm}  
    \subfloat[]{\includegraphics[width=4cm, height=3cm]{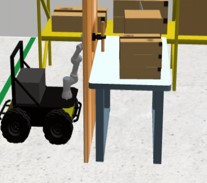}}  
    \caption{Various snapshots of a narrow passage motion planning using L\'evy flight based PRM. A UR3 manipulator is considered for simulation.}
\label{Intro_Figure}
\end{figure}

This paper is organized as follows. Section \ref{literature} presents a brief overview of the existing work in the field of narrow passage sampling. Section \ref{algorithm} presents the L\'evy flight based sampling scheme. Section \ref{simulation} and \ref{discussion} presents the simulation results and comparison with other narrow passage sampling schemes. Finally in Section \ref{conclusion} we conclude and outline possible extensions.

\section{Literature on Narrow Passage Sampling}\label{literature}
Narrow passage sampling methods can be divided into two groups - (a) path planning algorithms that can generate path through narrow passages intrinsically, (b) specific narrow passage sampling techniques that augment an existing path planning algorithm, such as PRM, by sampling configurations inside narrow regions. 

Among the first group, for example, \cite{flavigne2010improving},  combines the idea of Visibility-PRM\cite{simeon2000visibility} with multiple RRT trees. In \cite{nowakiewicz2010mst}, an MST is created with weighted $C_{space}$ cells, where weights depend on distance to boundary. In \cite{jaillet2011eg}, an environment-guided variant of RRT is designed for kinodynamic robot systems that estimates the probability of collision under uncertainty in control and sensing. In \cite{shkolnik2011sample}, spherical volumes are used instead of points for nodes in RRT tree expansion, resulting in formation of a sparse tree. \cite{vahrenkamp2011rdt+} uses dynamic extension of Bi-directional Rapidly exploring Dense Trees (RDT) \cite{lavalle2006planning}, by locally adapting the extend parameter for each node in narrow passage situations. In \cite{vonasek2011sampling}, the sampling for RRT is done along a pre-determined guiding path, with iterative scaling of the robot model in high dimension $C_{space}$. \cite{wedge2011using} generates local trees near obstacles from nodes, by assigning them a cost. In Toggle-PRM \cite{denny2011toggle}, both free and obstacle space are mapped, which increases sampling probability in narrow passages. In \cite{denny2013adapting}, a two level RRT expansion is proposed. In first level, the exploration area is identified based on node visibility and the tree adapts its expansion in case of narrow passage. In \cite{lee2012sr}, a bridge-line test and a non-colliding line test can identify regions around narrow passages. \cite{yeh2014umaprm} generates samples uniformly on the medial axis of $C_{free}$, combining the techniques of Medial Axis-PRM and UOBPRM. In \cite{wang2018learning}, learning based Multi-RRT models the tree selection process as a multi-armed bandit problem and uses Reinforcement Learning for action values with an improved $\epsilon$-greedy strategy. RRV in \cite{tahirovic2018rapidly} computes PCA to identify narrow passage entry or interior areas.

Among the second group, the most prominently used among the sampling strategies today are the Gaussian Sampler\cite{boor1999gaussian} and the Randomized Bridge Builder (RBB) Sampler \cite{hsu2003bridge} or the bridge test. The Gaussian Sampling strategy creates sampling points around obstacle surfaces, out of which some lie in the narrow region between two or more obstacles. In the RBB method, two random points inside the obstacle are extracted and connected and its midpoint, if in $C_{free}$, it is taken as a sample point. Many narrow passage sampling strategies have evolved from these, such as \cite{wang2010triple}, \cite{wang2010adaptive}, \cite{zhong2010robot} \& \cite{polden2013path}. In \cite{li2012extended}, sampled points using methods from \cite{zhong2010robot} are used with RRT-Connect\cite{kuffner2000rrt}.  In \cite{liu2012capacitor}, narrow passage samples are generated by building capacitor bridges between positive and negative toggled points in \textit{C-space} by flagging boundaries of moving obstacles. In \cite{yeh2012uobprm}, fixed length line segments are uniformly distributed in \textit{C-space} and their intersection points with \textit{C-obstacle} surfaces as retained as roadmap nodes. In \cite{denny2013lazy}, the authors combine \cite{denny2011toggle} with Lazy-PRM \cite{denny2013lazy} to show good sampling performance in narrow passages. Spark-PRM in \cite{shi2014spark} grows an RRT tree from a narrow passage sample generated from PRM, until a terminating condition is met and the tree is added to the roadmap. In \cite{li2014k}, groups of roadmaps are constructed and information about regions (whether near obstacle or in narrow passage) is stored in vertices of the group. In \cite{kang2016adaptive}, Adaptive Region Boosting and biased entropy are proposed to sample difficult areas, capturing temporal and spatial information of \textit{C-space}.

In the present paper, we select only RBB, Gaussian and Random walk to Surface (RWS) sampling mechanism (as these methods are foundational) to compare the performance with the proposed method. 

\section{L\'evy Flight based Narrow Passage Sampling}\label{algorithm}
\begin{figure}
\centering
\subfloat[\label{fig:levy.1}]{\includegraphics[width=0.5\linewidth,height=3cm,keepaspectratio=true]{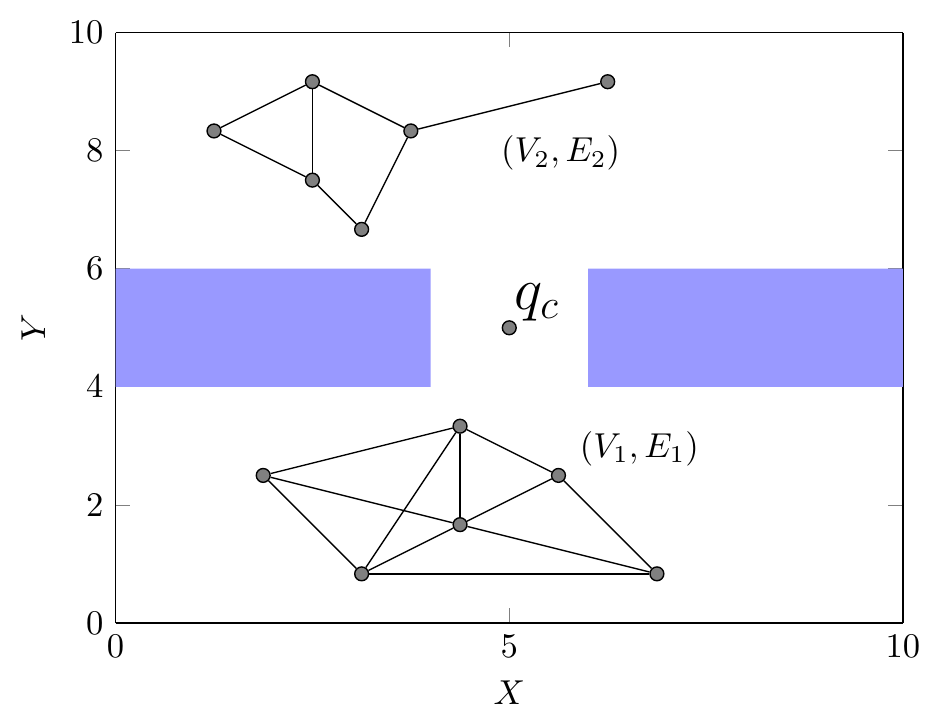}}
\subfloat[\label{fig:levy.2}]{\includegraphics[width=0.5\linewidth,height=3cm,keepaspectratio=true]{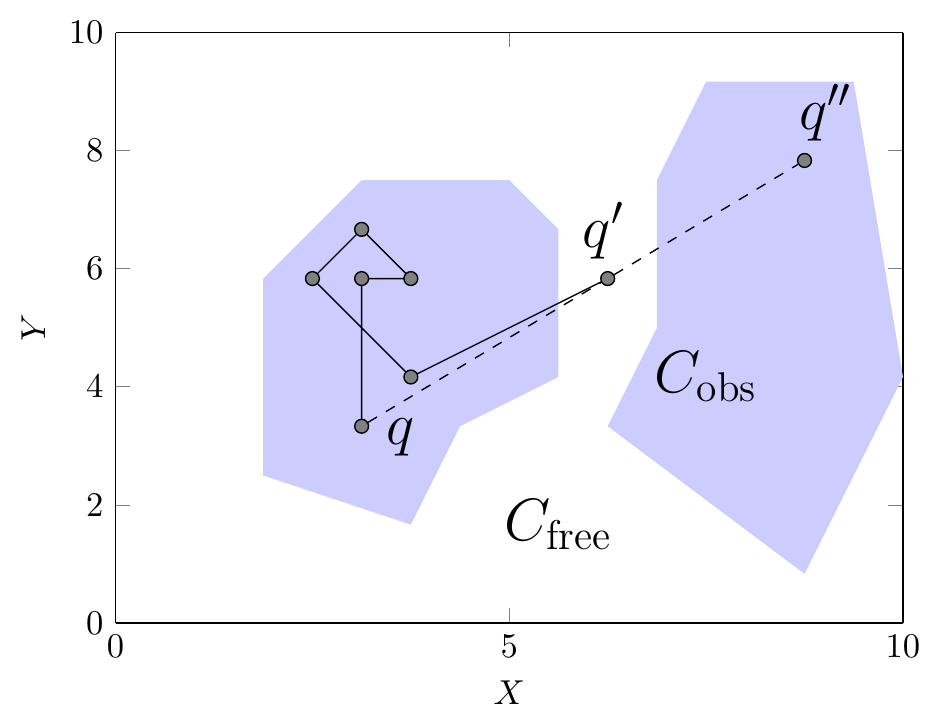}}
\caption{Schematics showing a) a critical configuration $q_c$ within a narrow passage may connect $2$ disconnected random path graph. b) Illustration of critical candidate samples ($q'$) generation using LFBS algorithm. A particle starts a L\'evy flight at $q$, until $q'$ followed by a bridge test check. }
\label{fig:levy}
\end{figure}

As shown in Fig.\ref{fig:levy.1}, the objective is to generate a critical configuration $q_c$ within the narrow passage of a configuration space to connect disconnected random graphs $(V_1,E_1)$ \& $(V_2,E_2)$. The proposed algorithm uses a basic random walk \cite{bera2010efficient}, which generates sample configurations on the obstacle boundaries. However, instead of performing the vanilla random walk, our approach utilizes a L\'evy flight mechanism (Fig.\ref{fig:levy.2}). 

\subsection{L\'evy Flight}
A L\'evy flight is a category of random walk in which the random walk step-size has a L\'evy distribution, a probability distribution that is heavy-tailed. The PDF of the length $X$ of each step is a power law function of the form
\begin{equation}
    f_X(x) \sim x^{-\alpha}
\end{equation}
Note that for $0 < \alpha <2$, $E[X] = \infty$, and $E[X^2] = \infty$, that is, infinite mean and variance. A higher value of $\alpha$ implies small steps and occasional long jumps which makes it difficult for a random walker to return to the starting point. That is, the walker can occasionally jump from one micro-structure to another which leads to a balanced exploration-exploitation scenario. In other words, the walk is intrinsically exploratory and efficient. For this reason, several other meta-heuristics such as PSO \cite{hariya2015levy}, Cuckoo search \cite{yang2009cuckoo} shows improved performances when a L\'evy flight based exploration technique is used. For details see \cite{shlesinger1995levy}.
\subsection{L\'evy Flight Sampler}
The proposed L\'evy Flight Sampler (LFS) algorithm makes use of the fact that from a sample configuration generated within an obstacle, one can come out of the obstacle by performing a L\'evy flight with step-size $l$. Algorithm \ref{algo1} represents the method to execute such a L\'evy Flight Sampler. To start the  L\'evy flight, the algorithm first checks for a random configuration $q$ to be colliding with an obstacle.  It then makes use of a \textit{L\'evy Flight} function to generate the position of new configuration $q'$ from $q$. That is, with a slight abuse of notation
\begin{equation}
q' \leftarrow q + (a + f(f_L(l),\theta'))
\end{equation} 
where $a$ is a base step-size (as used in \cite{bera2010efficient} and as bridge length in RBB). In $q'$, the angles for link (2D) or orientation (3D) for each configuration are obtained by 
\begin{equation}
\theta' = 2\pi(1 - e^{f_{\theta}(\theta)}) 
\end{equation} If the configuration $q'$ lies in $C_{free}$, it is regarded as a valid sample, or else the L\'evy flight continues. We choose $\alpha = 1.9$ for $f_L(l)$ and $\alpha = 1.1$ for $f_{\theta}(\theta)$.
\begin{algorithm} 
\small{}
\SetAlgoLined
 \For{$ i \gets 1 \ to \ K $ }{
  $ q \gets \ Random \ Configuration$;\\
  \If{$ q \in C_{obs}$}
  {
   Flag = 0\;
   \While{Flag = 0}
   {
   	Select Random Direction\;
   	$ q' \gets \ L\acute{e}vy \ Flight (q,l)$\;
   	\eIf{$ q' \in C_{free}$}
   	{
   		add $q'$ to the list $L$\;
   		Flag = 1;
   	}{
   $ q \gets q'$\;
  }
  }
  }
 }
 Return L;
 \caption{L\'evy Flight Sampler}
 \label{algo1}
\end{algorithm}

\subsection{L\'evy Flight Bridge Sampler}
The L\'evy Flight Bridge Sampler (LFBS) is an extension of the above proposed LFS. In this, once $q'$ is obtained in the free configuration space, $C_{free}$, an additional check is performed to ensure that $q'$ is in the narrow passage indeed. This check is performed using $extend$ function as defined in Algorithm \ref{algo2}. For comparison purposes, a similar bridge sampler with only basic random walk also considered and outlined in the following.
\begin{algorithm} 
\small{}
\SetAlgoLined
 \For{$ i \gets 1 \ to \ K $ }{
  $ q \gets \ Random \ Configuration$;\\
  \If{$ q \in C_{obs}$}
  {
   Flag = 0\;
   \While{Flag = 0}
   {
   	Select Random Direction\;
   	$ q' \gets \ L\acute{e}vy \ Flight (q,l)$\;
   	\eIf{$ q' \in C_{free}$}
   	{
   		$ q'' \gets extend(q,q')$\;
   		\If{$ q'' \in C_{obs}$}
   		{
   			add $q'$ to the list $L$\;
   			Flag = 1;
   		}
   	}{
   $ q \gets q'$\;
  }
  }}
  
 }
 Return L\; 
  
 \SetKwProg{Def}{def}{:}{}
\Def{extend($q$,$q'$)}
	{
		extend the line joining $q$ and $q'$ to $q''$\;
		such that $q'$ is mid-point of $qq''$\;
		Return $q''$\;
	}
 \caption{L\'evy Flight Bridge Sampler}
 \label{algo2}
\end{algorithm}

\subsection{Random Walk Bridge Sampler}
The Random Walk Bridge Sampler (RWBS) is a modification of the RWS \cite{bera2010efficient} and follows similar sample generation mechanism of LFBS, except the L\'evy flight is replaced by a conventional random walk. The $extend$ function (of Algorithm \ref{algo2}) is used on $C_{free}$ sampled configurations to confirm the narrow passage.

\section{Simulation Results}\label{simulation} 
In this section, the performance of the above proposed narrow passage sampling methods, that is, Levy Flight Sampler (LFS), Randowm Walk Bridge Sampler (RWBS) and L\'evy Flight Bridge Sampler (LFBS), are compared with the standard Gaussian sampler, Randomised Bridge Builder Sampler (RBBS) and Random Walk Sampler (RWS). The comparison has been done by extensively simulating narrow passage scenarios in 2-D as well as a 3-D environments. In 2-D environment, experiments are conducted for a $7$-DOF manipulator, and simulations are conducted considering a Bar shape and a Joint shape obstacle scenarios. Note that, although the workspace is in 2-D, the corresponding configuration space is $\mathbb{R}^2 \times S^7$ which are locally like $\mathbb{R}^9$. In case of 3-D environment, an L-shape rigid robot is considered in a Joint Shape and a Teeth Shape obstacle scenario to obtain the comparison results. The L\'evy flight based method and the RBBS are also integrated in PRM planner to test and compare the quality of motion planning for a $6$-DOF UR3 manipulator mounted on a ground vehicle and a Quad-copter based inspection of 1) a tunnel, and 2) a maze like scenario. 

% \begin{figure} 
%     \centering
%     \subfloat[LFS\label{fig:fig1B.1}]{\includegraphics[width=3cm, height=2cm]{figures/Fig2D_Bar_5DOF_30points_10step_lrws.png}}  
%     \subfloat[RWS\label{fig:fig1B.2}]{\includegraphics[width=3cm, height=2cm]{figures/Fig2D_Bar_5DOF_30points_10step_rws.png}}
%     \subfloat[Gaussian\label{fig:fig1B.3}]{\includegraphics[width=3cm, height=2cm]{figures/Fig2D_Bar_5DOF_30points_10step_gaussian.png}}\\
%     \subfloat[RBBS\label{fig:fig1B.4}]{\includegraphics[width=3cm, height=2cm]{figures/Fig2D_Bar_5DOF_30points_10step_rbbs.png}}
%     \subfloat[RWBS\label{fig:fig1B.5}]{\includegraphics[width=3cm, height=2cm]{figures/Fig2D_Bar_5DOF_30points_10step_rwbs.png}}
%     \subfloat[LFBS\label{fig:fig1B.6}]{\includegraphics[width=3cm, height=2cm]{figures/Fig2D_Bar_5DOF_30points_10step_lrwbs.png}}
%     \caption{Narrow passage samples in a 2D environment for a $5$-Link Robot with a Bar Shape obstacle. Step-size 10 units}
% \label{fig:fig1B}
% \end{figure}
\begin{figure}  
    \centering
    \subfloat[LFS\label{fig:fig1C.1}]{\includegraphics[width=3cm, height=2cm]{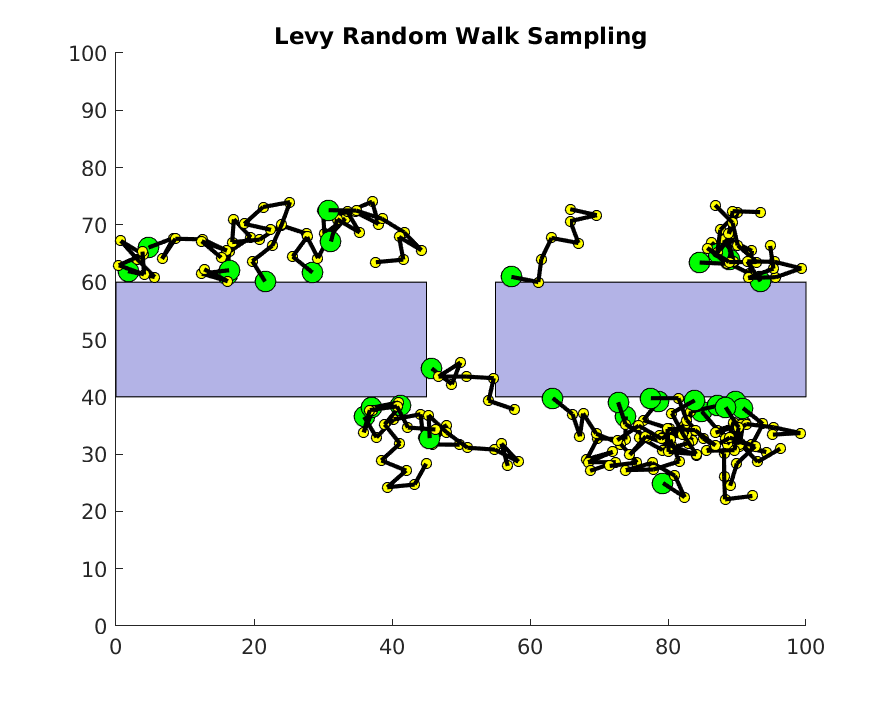}}  
    \subfloat[RWS\label{fig:fig1C.2}]{\includegraphics[width=3cm, height=2cm]{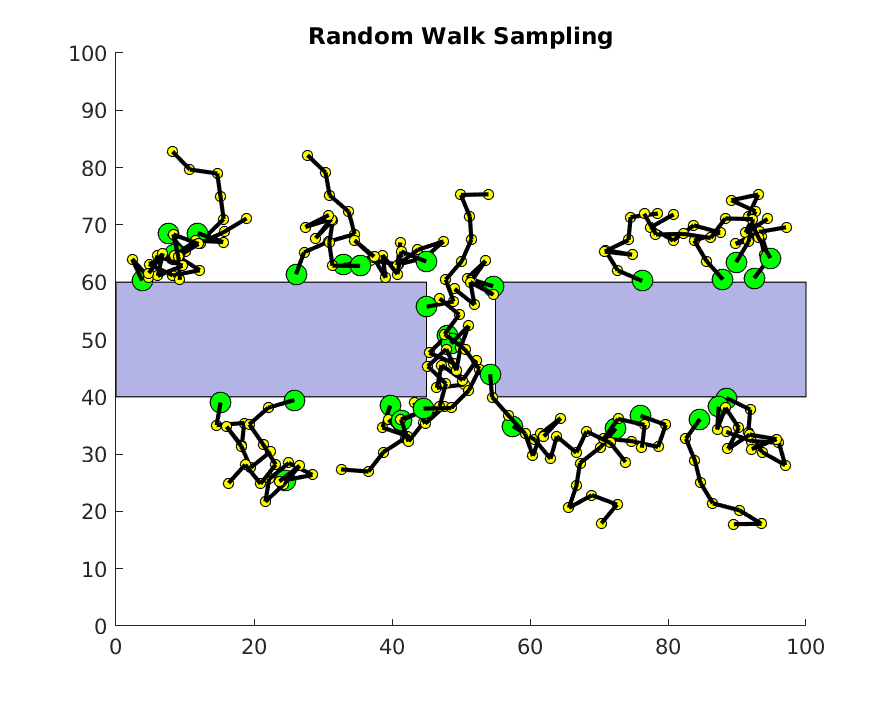}}
    \subfloat[Gaussian\label{fig:fig1C.3}]{\includegraphics[width=3cm, height=2cm]{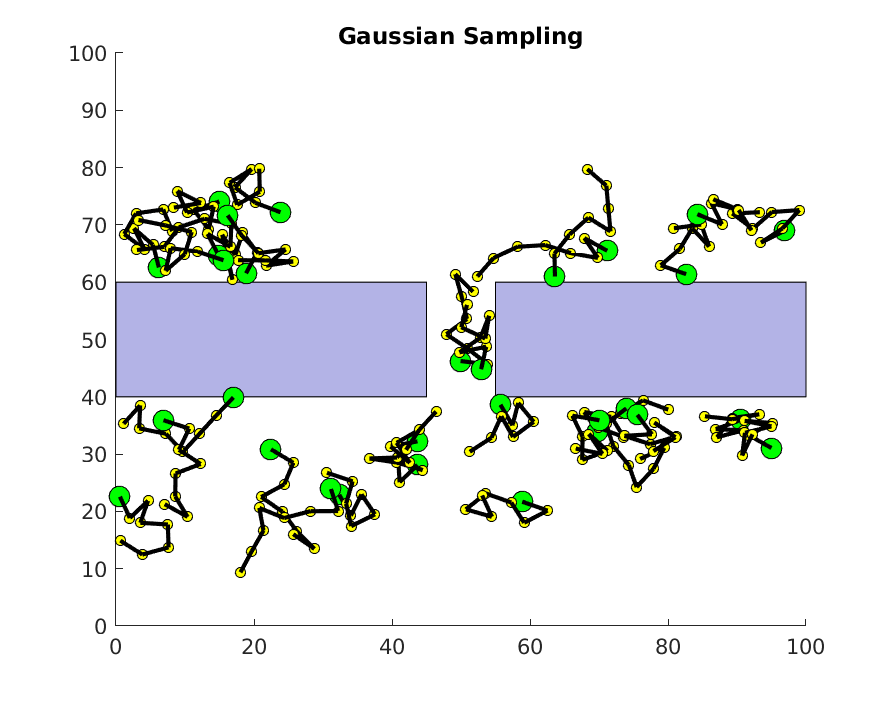}}\\
    \subfloat[RBBS\label{fig:fig1C.4}]{\includegraphics[width=3cm, height=2cm]{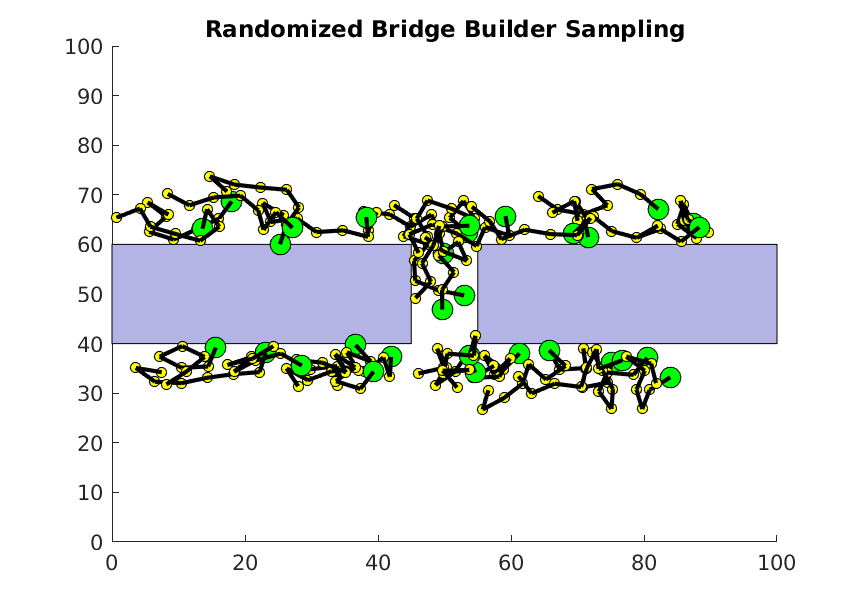}}
    \subfloat[RWBS\label{fig:fig1C.5}]{\includegraphics[width=3cm, height=2cm]{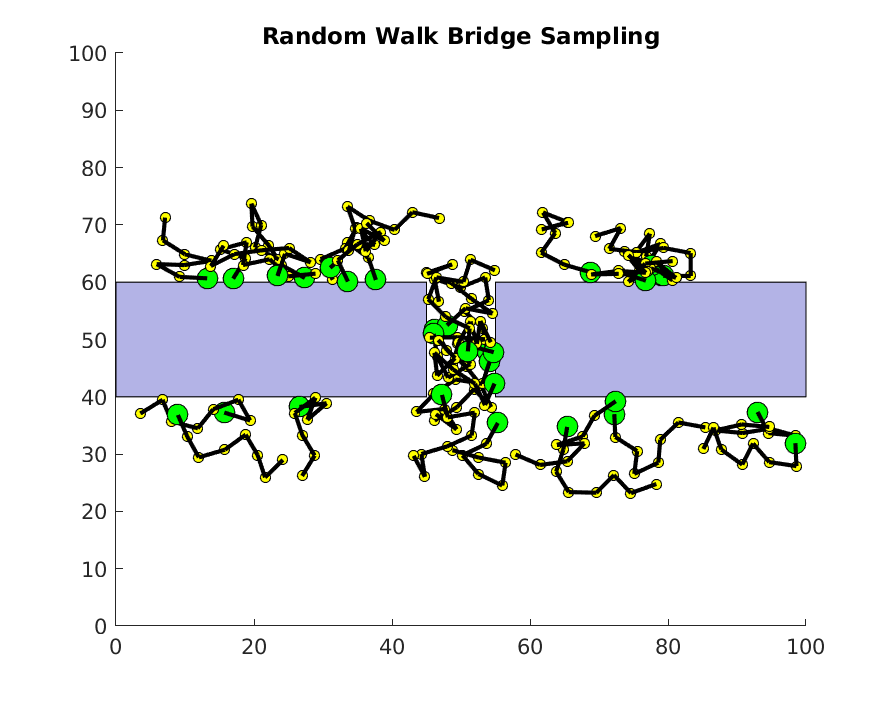}}
    \subfloat[LFBS\label{fig:fig1C.6}]{\includegraphics[width=3cm, height=2cm]{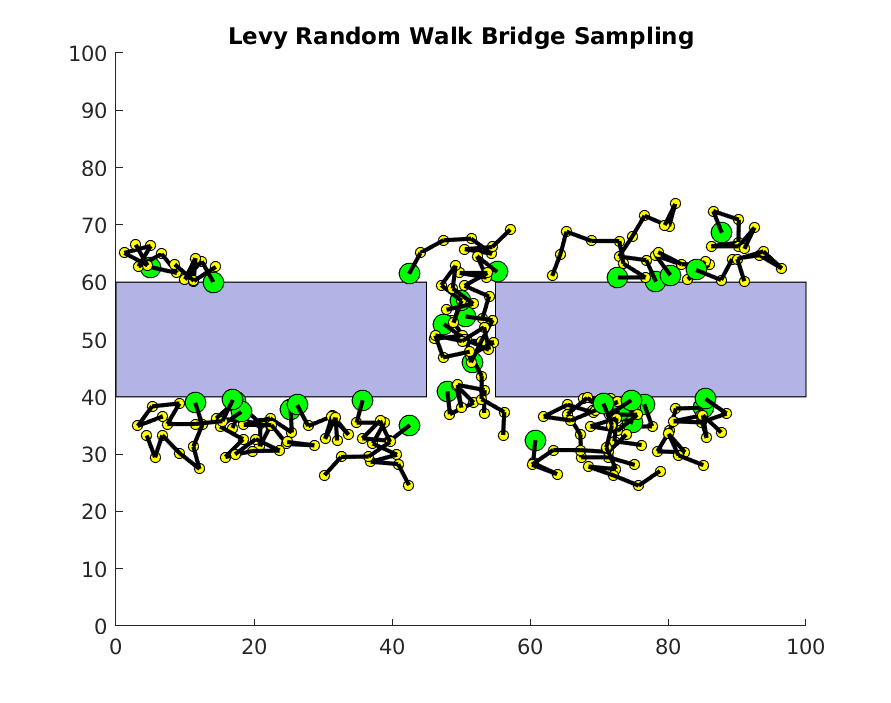}}
    \caption{Narrow passage samples in a 2D environment for a $7$-DOF Robot with a Bar Shape obstacle. Step-size 2 units}
\label{fig:fig1C}
\end{figure}
% 1.) Collision for Multi-DoF Robot in Bar Shape obstacle scenario
\begin{figure} 
    \centering
    % \subfloat[5-Link\label{fig:graph1.2}]{\includegraphics[width=0.46\columnwidth, height=2cm,keepaspectratio=false]{5Link_Bar_Collision.pdf}}
    \subfloat[Time\label{fig:graph2.3}]{\includegraphics[width=0.46\columnwidth, height=2cm,keepaspectratio=false]{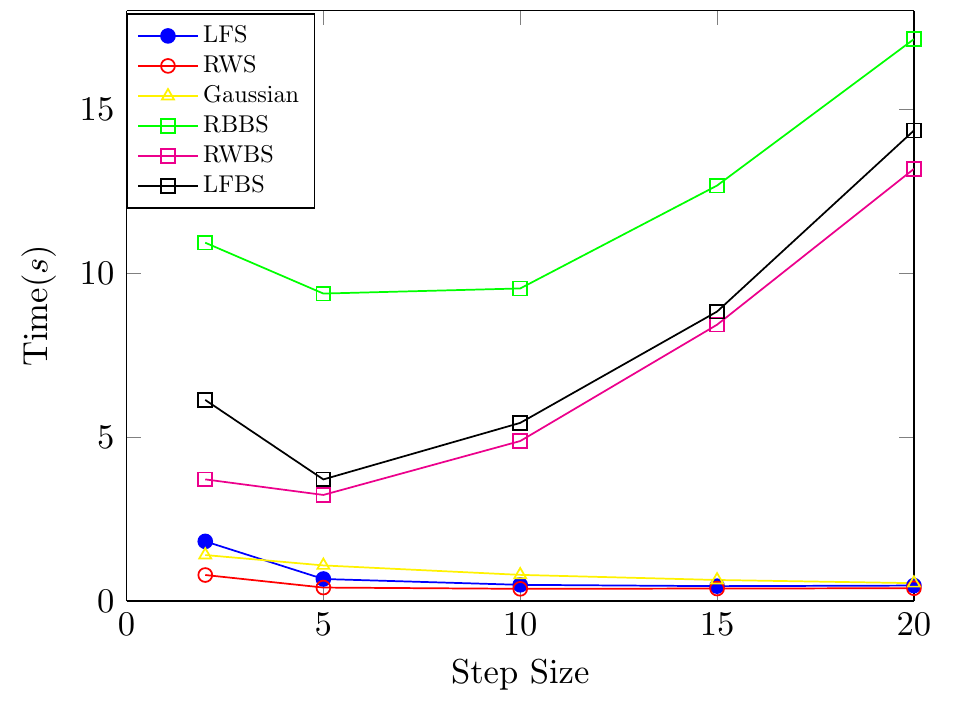}}
    \subfloat[No. of Collisions\label{fig:graph1.3}]{\includegraphics[width=0.46\columnwidth, height=2cm,keepaspectratio=false]{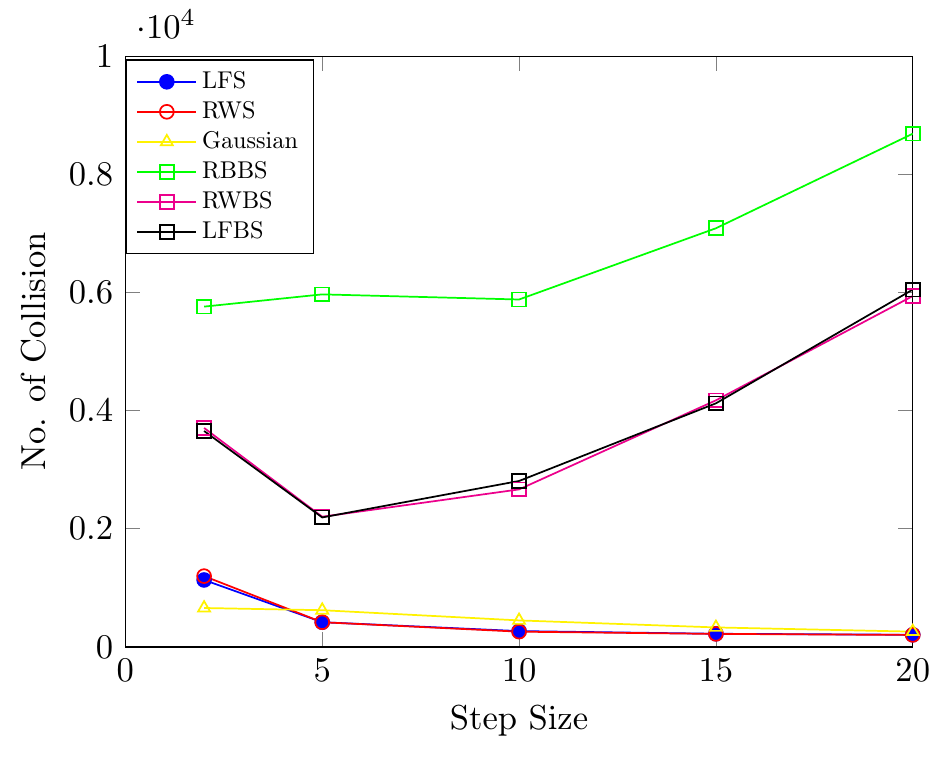}}
    \caption{Collision calls and execution time comparisons for $7$-DoF Robot in Bar shaped obstacle scenario}
    \label{fig:graph1}
\end{figure}
% % % 2.) Time for Multi-DoF Robot in Bar Shape obstacle scenario
% \begin{figure}  
%     \centering
%     \subfloat[5-Link\label{fig:graph2.2}]{\includegraphics[width=0.46\columnwidth, height=2cm,keepaspectratio=false]{5Link_Bar_Time.pdf}}
%     \subfloat[7-Link\label{fig:graph2.3}]{\includegraphics[width=0.46\columnwidth, height=2cm,keepaspectratio=false]{7Link_Bar_Time.pdf}}
%     \caption{Execution time comparison for Multi-DoF Robot in Bar Shape obstacle scenario}
%     \label{fig:graph2}
% \end{figure}

The environment considered for narrow passage sampling simulation is a $100$ unit sided square area and cube volume in 2-D and 3-D respectively. The width of the narrow passage between the obstacles varies between $5$ to $10$ units depending on the obstacle scenario. For comparison, the total execution time $t_\varepsilon$ and number of collision checking calls (as collision checking calls are expensive), $\gamma_c$, to generate a fixed number of candidate samples, $K_c$, here $K_c = 30$, are considered. To have a comprehensive comparison of all the above mentioned parameters, all the algorithms are run in the simulated environment with step-size, $a$, varying from 2 units to 20 units. All simulations are conducted on a system with Intel Core i7-7500U CPU @ 2.70GHz * 4 processor and 8GB of RAM. Algorithms are executed for $300$ times for each obstacle scenario and for each \textit{step-size}, to approximate the average case performance.

In the first experiment, candidate samples are generated in the narrow region for the Bar Shape obstacle scenario, with different step size $a \in \{2,10\}$ and for a $7$-DOF manipulator, (as shown in Fig. \ref{fig:fig1C}). The corresponding $t_\varepsilon$ and $\gamma_c$ comparison results are shown in Fig. \ref{fig:graph1}. It is evident that both the parameters are less for the proposed LFBS and RWBS when compared to RBBS. Even though the parameters $t_\varepsilon$, and $\gamma_c$ are very less for Gaussian, RWS and LFS, visual inspection shows that the sample quality (Fig. \ref{fig:fig1C}) is very poor. That is, most samples are not in the critical zone. The sampling quality of LWBS and RWBS can be seen as comparable with that of RBBS. 

\begin{figure} 
    \centering
    \subfloat[LFS\label{fig:fig2C.1}]{\includegraphics[width=3cm, height=2.0cm]{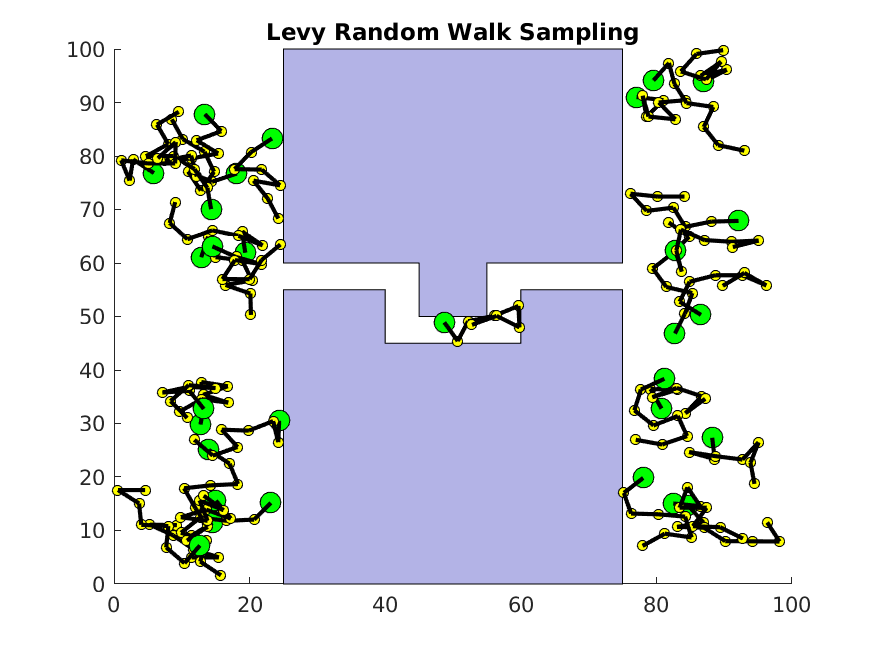}}  
    \subfloat[RWS\label{fig:fig2C.2}]{\includegraphics[width=3cm, height=2.0cm]{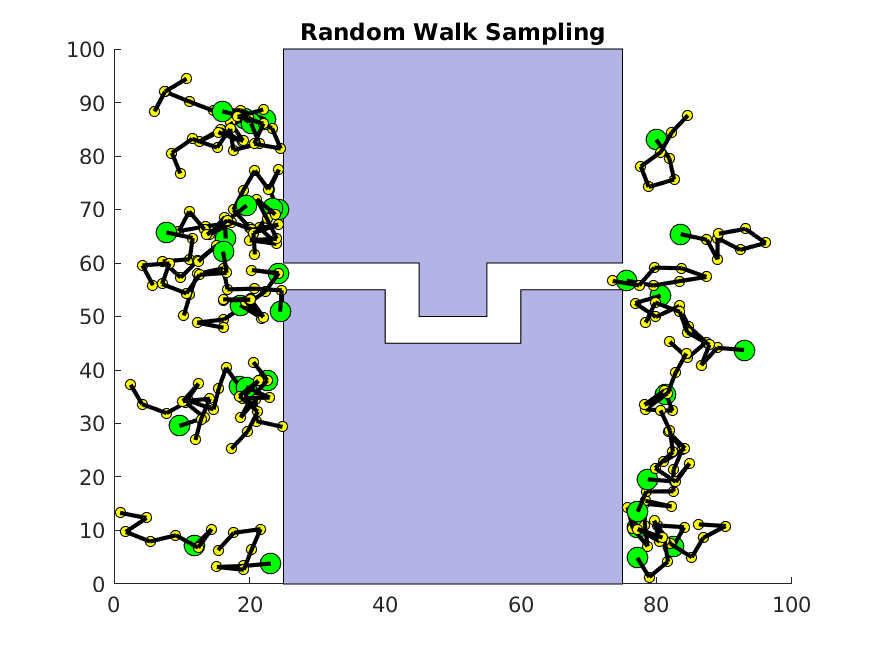}}
    \subfloat[Gaussian\label{fig:fig2C.3}]{\includegraphics[width=3cm, height=2.0cm]{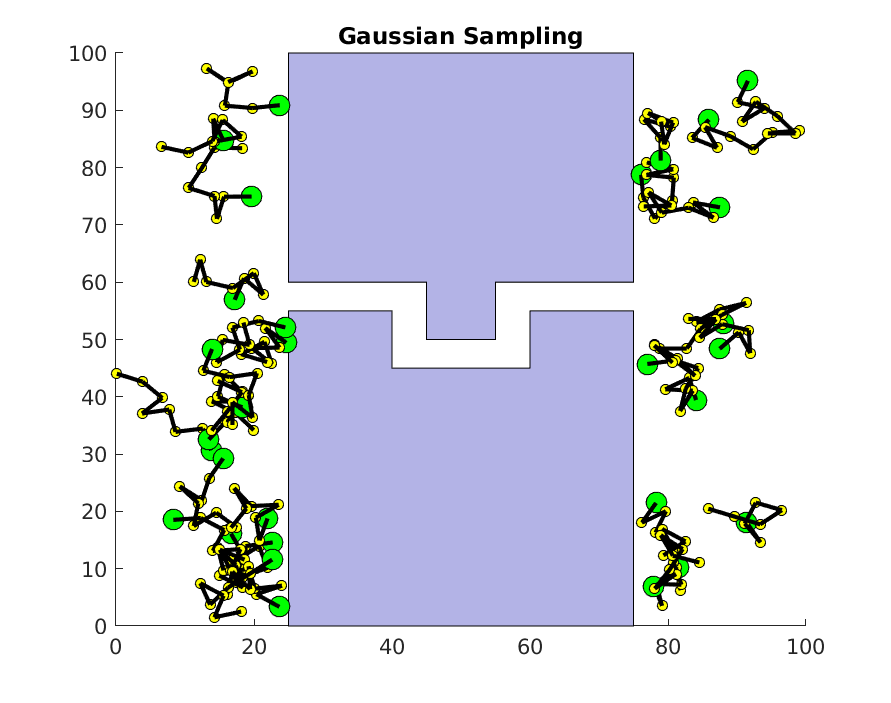}}\\
    \subfloat[RBBS\label{fig:fig2C.4}]{\includegraphics[width=3cm, height=2.0cm]{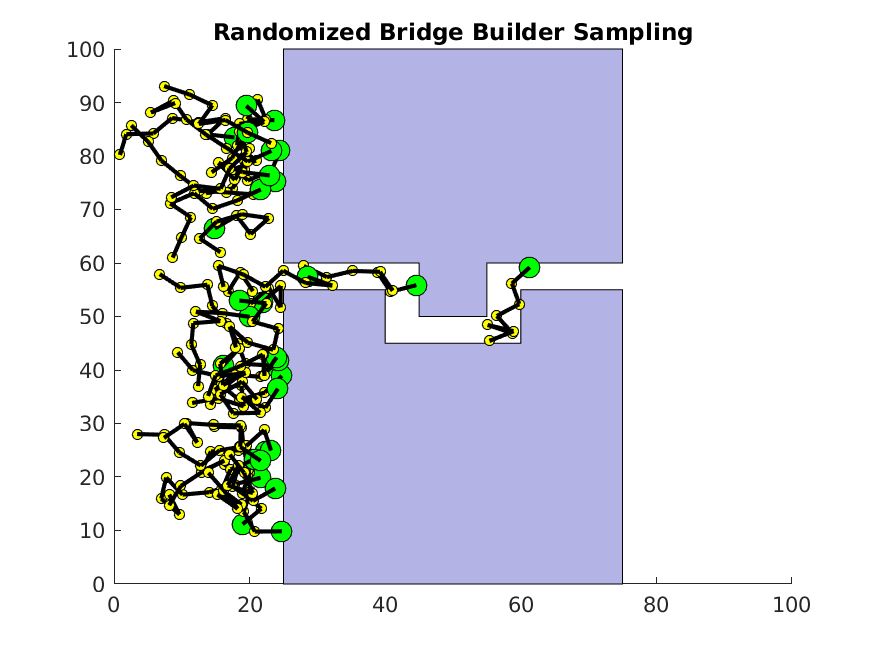}}
    \subfloat[RWBS\label{fig:fig2C.5}]{\includegraphics[width=3cm, height=2.0cm]{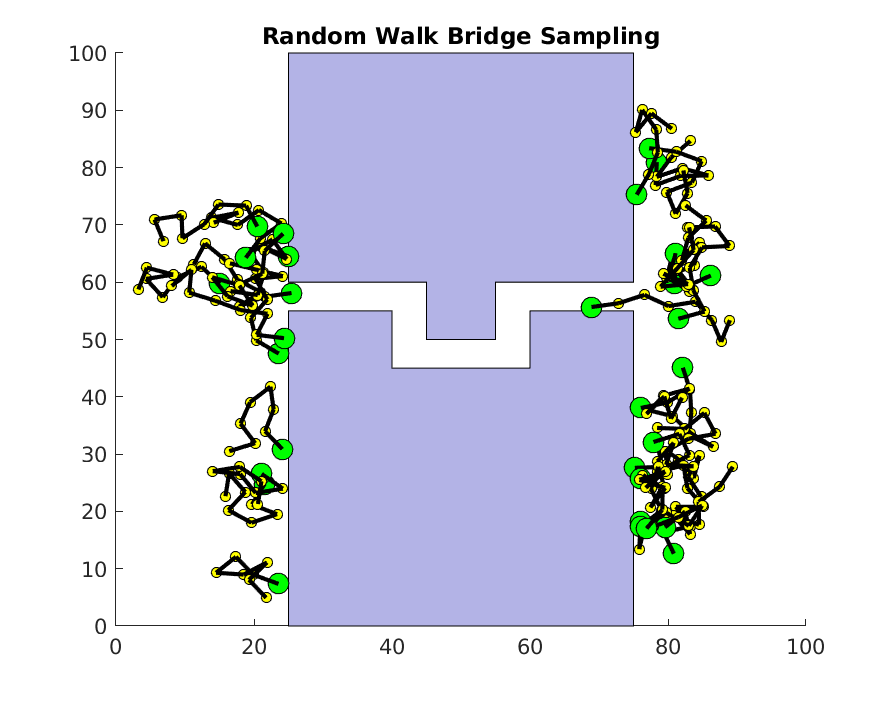}}
    \subfloat[LFBS\label{fig:fig2C.6}]{\includegraphics[width=3cm, height=2.0cm]{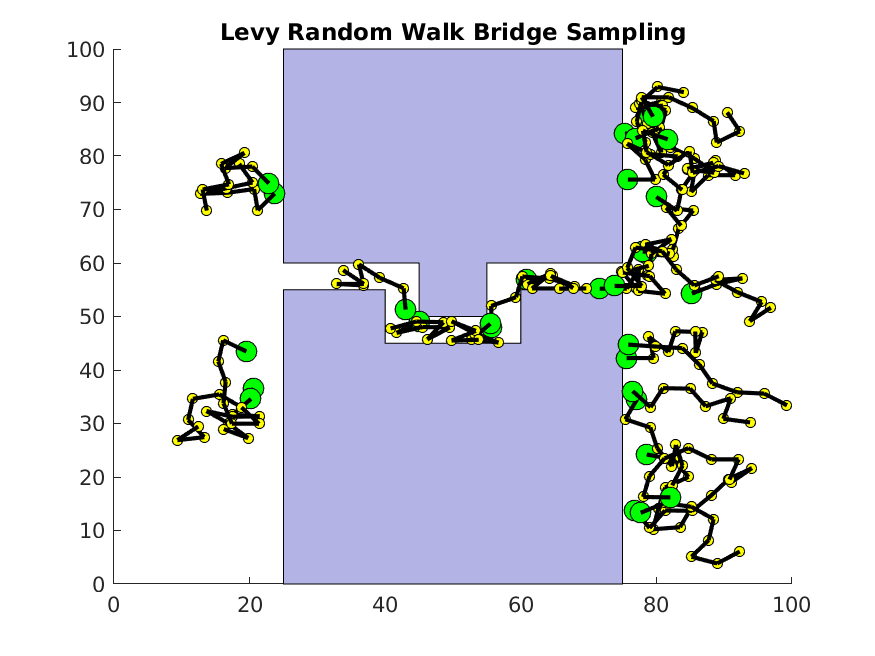}}
    \caption{Narrow passage samples in a 2D environment for a 7-DOF Robot in Joint Shape obstacle scenario for step-size 10 units}
\label{fig:fig2C}
\end{figure}

\begin{figure} 
    \centering
    % \subfloat[5-Link\label{fig:graph3.2}]{\includegraphics[width=0.46\columnwidth, height=2cm,keepaspectratio=false]{5Link_Joint_Collision.pdf}}
    \subfloat[Time\label{fig:graph4.3}]{\includegraphics[width=0.46\columnwidth, height=2cm,keepaspectratio=false]{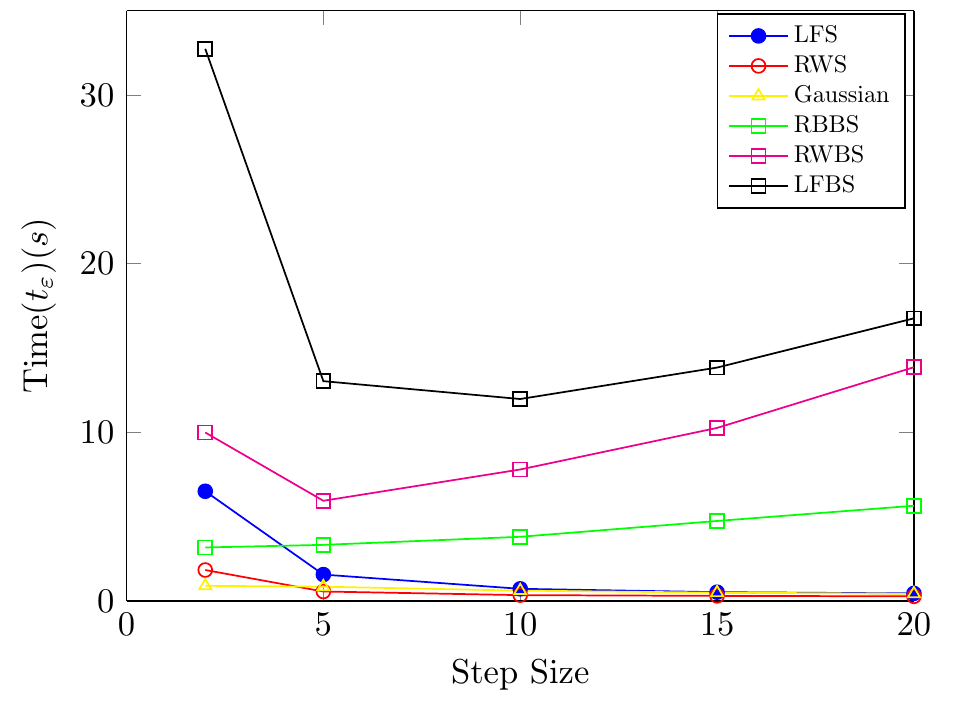}}
    \subfloat[No. of Collisions\label{fig:graph3.3}]{\includegraphics[width=0.46\columnwidth, height=2cm,keepaspectratio=false]{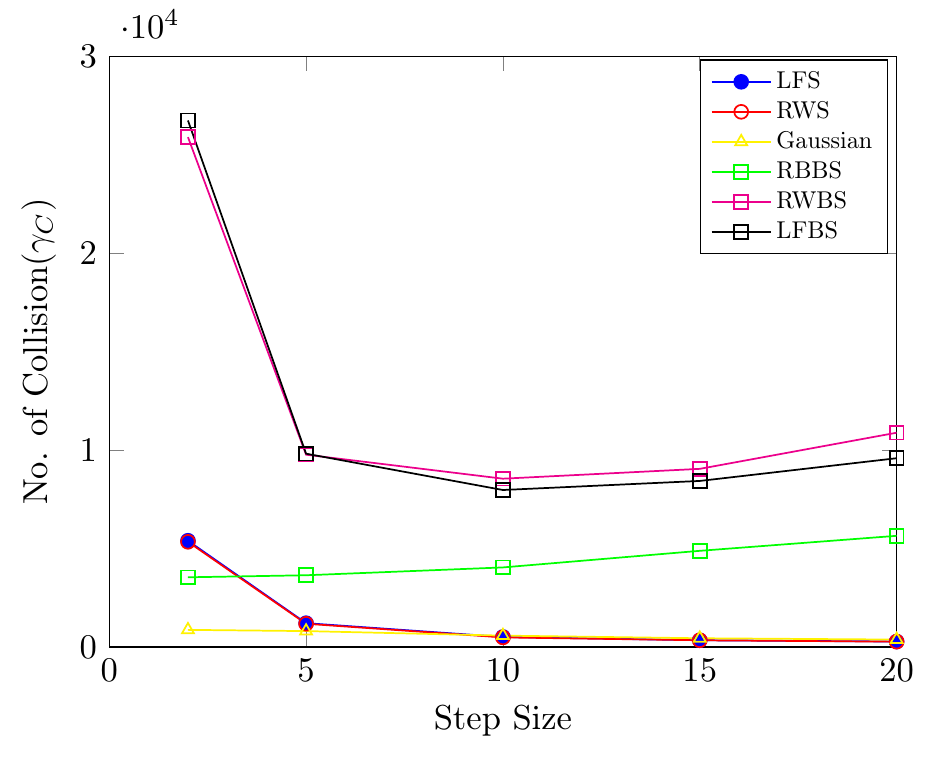}}
    \caption{Collision calls and execution time comparisons for $7$-DoF Robot in Joint Shape obstacle scenario}
\label{fig:graph3}
\end{figure}

For Joint Shape obstacle scenario (Fig. \ref{fig:fig2C}), the results are shown in Fig. \ref{fig:graph3}. The parameters $t_\varepsilon$, and $\gamma_c$ here are comparable for a wide range of step-sizes, $a$, for all the algorithms except the LFBS and RWBS, which show more \textit{time/collision checks} than others. However, Fig. \ref{fig:fig2C} reveals that the sample quality for LFBS is far superior than the other strategies for both the same case. 

% 3.) Collision for Multi-DoF Robot in Joint Shape obstacle scenario
%4.) Time for Multi-DoF Robot in Joint Shape obstacle scenario
% \begin{figure} 
%     \centering
%     \subfloat[5-Link\label{fig:graph4.2}]{\includegraphics[width=0.46\columnwidth, height=2cm,keepaspectratio=false]{5Link_Joint_Time.pdf}}
%     \subfloat[7-Link\label{fig:graph4.3}]{\includegraphics[width=0.46\columnwidth, height=2cm,keepaspectratio=false]{7Link_Joint_Time.pdf}}
%   \caption{Execution time for Multi-DoF Robot in Joint Shape obstacle scenario}
% \label{fig:graph4}
% \end{figure}

\begin{figure} 
    \centering
    \subfloat[LFS\label{fig:fig4.1}]{\includegraphics[width=3cm, height=2cm]{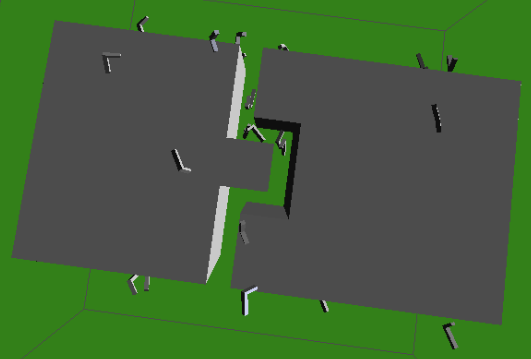}}  
    \subfloat[RWS\label{fig:fig4.2}]{\includegraphics[width=3cm, height=2cm]{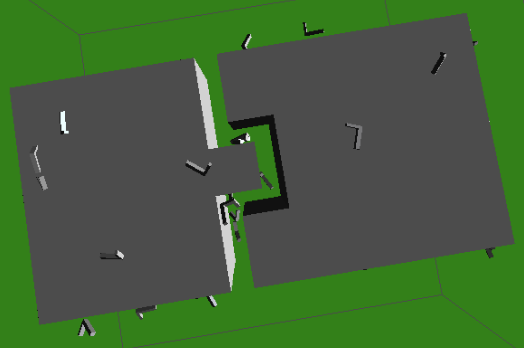}}
    \subfloat[Gaussian\label{fig:fig4.3}]{\includegraphics[width=3cm, height=2cm]{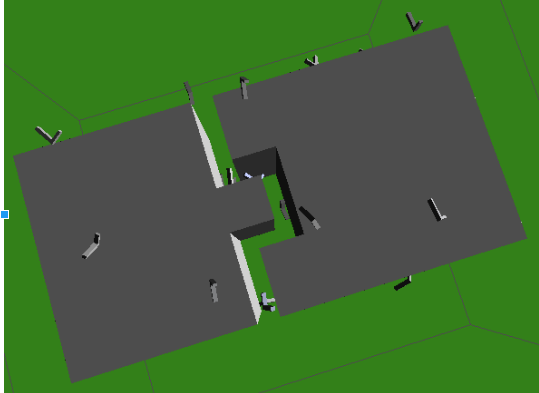}}\\
    \subfloat[RBBS\label{fig:fig4.4}]{\includegraphics[width=3cm, height=2cm]{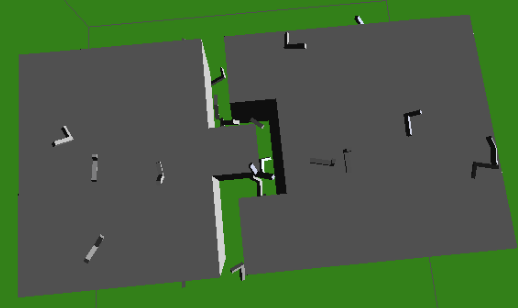}}
    \subfloat[RWBS\label{fig:fig4.5}]{\includegraphics[width=3cm, height=2cm]{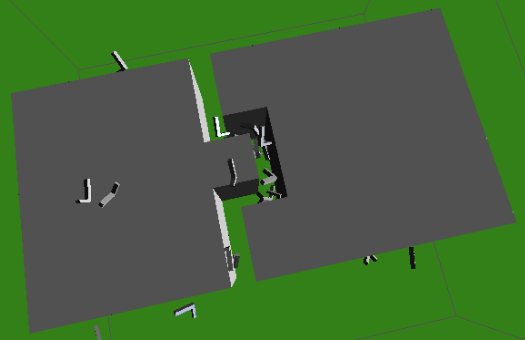}}
    \subfloat[LFBS\label{fig:fig4.6}]{\includegraphics[width=3cm, height=2cm]{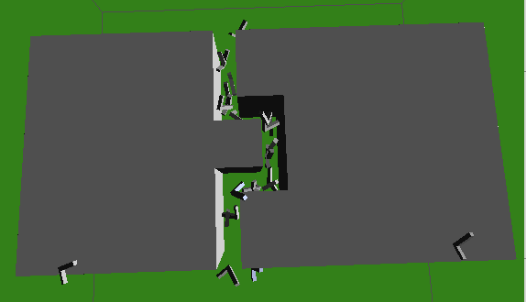}}
\caption{Narrow passage samples in a 3D environment for a Joint Shape obstacle scenario}
\label{fig:fig4}
\end{figure}

\begin{figure} 
    \centering
    \subfloat[No. of Collisions\label{fig:graph6.1}]{\includegraphics[width=0.46\columnwidth, height=2cm,keepaspectratio=false]{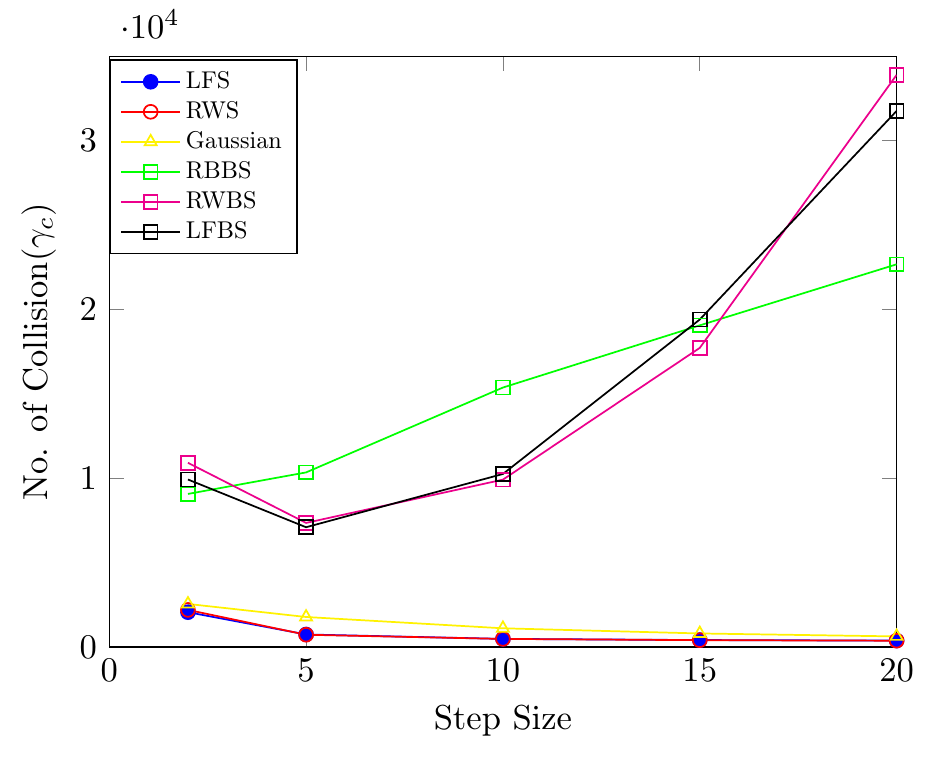}}
    \subfloat[Time\label{fig:graph6.2}]{\includegraphics[width=0.46\columnwidth, height=2cm,keepaspectratio=false]{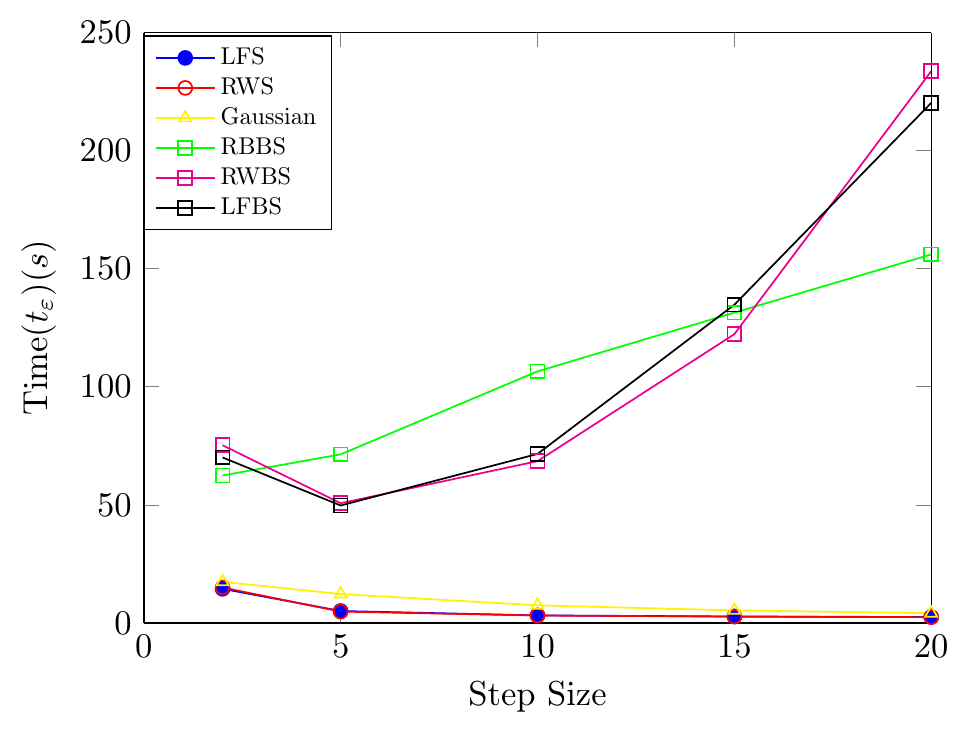}}
\caption{Execution time and collision calls in 3D in Joint Shape obstacle scenario}
\label{fig:graph6}
\end{figure}

\begin{figure} 
    \centering
    \subfloat[LFS\label{fig:fig5.1}]{\includegraphics[width=3cm, height=2cm]{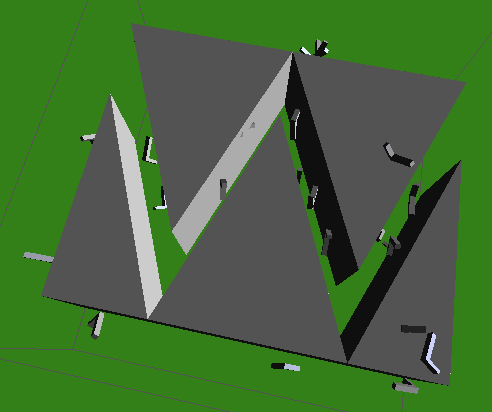}}  
    \subfloat[RWS\label{fig:fig5.2}]{\includegraphics[width=3cm, height=2cm]{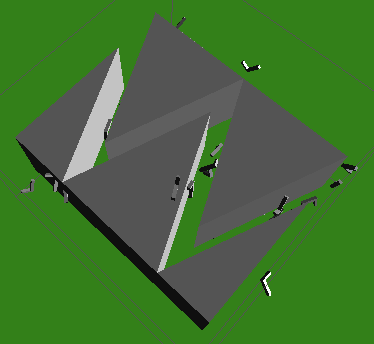}}
    \subfloat[Gaussian\label{fig:fig5.3}]{\includegraphics[width=3cm, height=2cm]{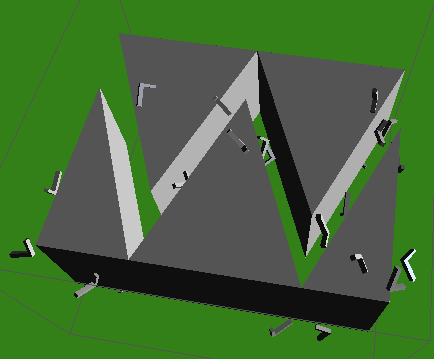}}\\
    \subfloat[RBBS\label{fig:fig5.4}]{\includegraphics[width=3cm, height=2cm]{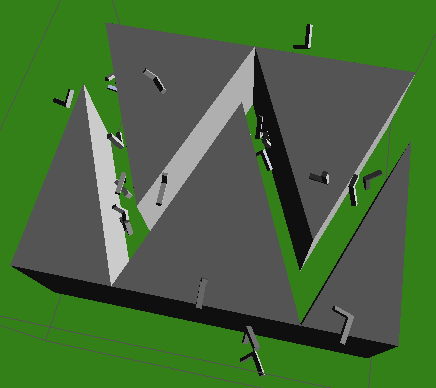}}
    \subfloat[RWBS\label{fig:fig5.5}]{\includegraphics[width=3cm, height=2cm]{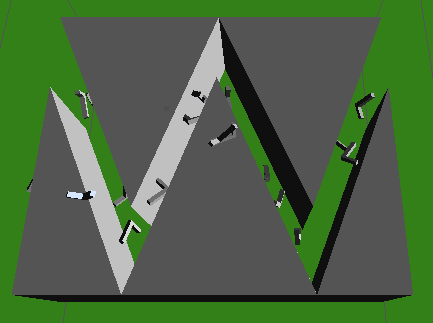}}
    \subfloat[LFBS\label{fig:fig5.6}]{\includegraphics[width=3cm, height=2cm]{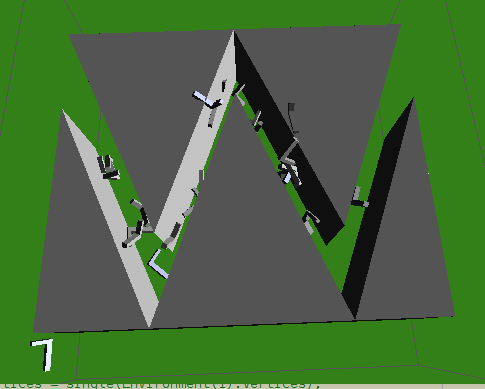}}
\caption{Narrow passage samples in a 3D environment for a Teeth Shape obstacle scenario}
\label{fig:fig5}
\end{figure}

\begin{figure} 
    \centering
    \subfloat[No. of Collisions\label{fig:graph7.1}]{\includegraphics[width=0.46\columnwidth, height=2cm,keepaspectratio=false]{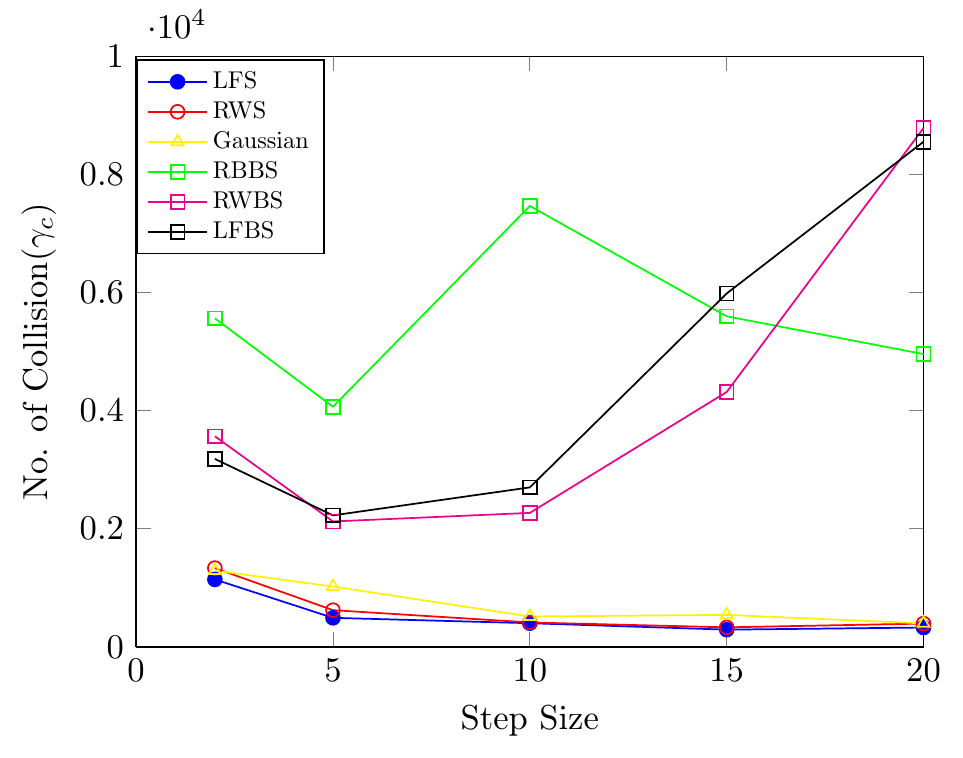}}
    \subfloat[Time\label{fig:graph7.2}]{\includegraphics[width=0.46\columnwidth, height=2cm,keepaspectratio=false]{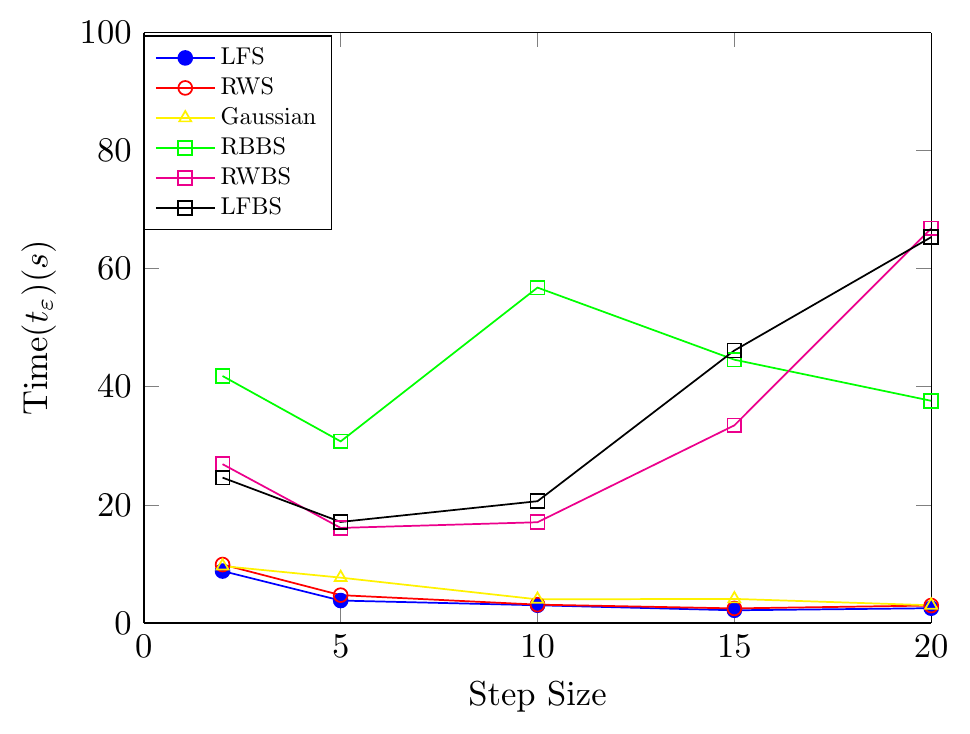}}
\caption{Execution time and collision calls in 3D in Teeth Shape obstacle scenario}
\label{fig:graph7}
\end{figure}

The critical samples generation results in a 3D-environment are shown in the following. For a Joint Shape obstacle scenario, as shown in Fig. (\ref{fig:fig4}, \ref{fig:graph6}), the $t_\varepsilon$, and $\gamma_c$ for LFBS and RWBS are less than or comparable to those of RBBS. Also, the sample quality of LFBS and RWBS is better, with more number of key configurations in the confined region (Fig. \ref{fig:fig4}). The corresponding result for Gaussian, RWS and LFS shows the samples are scattered around the obstacles and not just within the narrow region of the workspace. For the Teeth shaped obstacle scenario, results obtained show superiority of LFBS and RWBS in terms of \textit{time/collision checks} and sampling quality, as is evident from Fig. \ref{fig:fig5} and Fig. \ref{fig:graph7}.

Next we present the simulation results when the LFBS and RBBS methods are included in a PRM like planner.
% 6.) Time and collision in 3D in Joint Shape obstacle scenario
% 5.) Figure for Sampling Comparison in 3D in Teeth Shape obstacle scenario
% 7.) Time and collision in 3D in Teeth Shape obstacle scenario
% 6.) Figure for Sampling Comparison of Spitfire in 3D in Teeth Shape obstacle scenario
% \begin{figure} 
%     \centering
%     \subfloat[LFS\label{fig:fig5.1}]{\includegraphics[width=3cm, height=2cm]{figures/Fig_3D_Teeth_Spitfire_LWS.png}}  
%     \subfloat[RWS\label{fig:fig5.2}]{\includegraphics[width=3cm, height=2cm]{figures/Fig_3D_Teeth_Spitfire_RWS.png}}
%     \subfloat[Gaussian\label{fig:fig5.3}]{\includegraphics[width=3cm, height=2cm]{figures/Fig_3D_Teeth_Spitfire_Gaussian.png}}\\
%     \subfloat[RBBS\label{fig:fig5.4}]{\includegraphics[width=3cm, height=2cm]{figures/Fig_3D_Teeth_Spitfire_RBBS.png}}
%     \subfloat[RWBS\label{fig:fig5.5}]{\includegraphics[width=3cm, height=2cm]{figures/Fig_3D_Teeth_Spitfire_RWBS.png}}
%     \subfloat[LFBS\label{fig:fig5.6}]{\includegraphics[width=3cm, height=2cm]{figures/Fig_3D_Teeth_Spitfire_LWBS.png}}
% \caption{Narrow passage samples in a 3D environment for a UAV in a Teeth Shape obstacle scenario}
% \label{fig:fig6}
% \end{figure}
% 8.) Time and collision in 3D of Spitfire in Joint Shape obstacle scenario
% \begin{figure} 
%     \centering
%     \subfloat[No. of Collisions\label{fig:graph8.1}]{\includegraphics[width=0.46\columnwidth, height=2cm,keepaspectratio=false]{3DUAV_Collision.pdf}}
%     \subfloat[Time\label{fig:graph8.2}]{\includegraphics[width=0.46\columnwidth, height=2cm,keepaspectratio=false]{3DUAV_Time.pdf}}
% \caption{Execution time and collision calls in 3D for an UAV in a Joint Shape obstacle scenario}
% \label{fig:graph8}\end{figure}
\subsection{Critical Sampling based PRM Planner Evaluation}
We also tested and compared the trajectories, when such narrow passage sampling strategies (that is, RBBS and LFBS) are included in the sample generation module, of a PRM based UAV motion planning scenario for different environments, such as maze and tunnels. Additional experiments for a PRM based $6$-DOF UR3 manipulator motion planning is also carried out based on the said sampling strategies. All the simulations are performed in a ROS-Gazebo environment. For collision detection we use octomap of the robot's view using kinect camera and subsequently use FCL (Flexible Collision Library) \cite{pan2012fcl} to reject those states which are in collision with the environment. For manipulator motion planning simulations, various environments are considered such as, object manipulation \& placement in a constrained space, as shown in Fig. \ref{Intro_Figure}, and a steering for welding operations in a cluttered space, as shown in Fig. \ref{fig:chasis}.
  
% \begin{figure}[h]
%     \centering
%     {\includegraphics[width=0.98\columnwidth, height=4cm,keepaspectratio=true]{figures/landslide_final_1.png}}
% \caption{Quadrotor inspecting a building after landslide (environment size $70m\times70m\times15m$) }
% \label{fig:landslide}
% \end{figure}
\begin{figure}
    \centering
    \subfloat[$PRM_{LFBS}$\label{Maze_LFBS}]{\includegraphics[height= 4cm,width = 4cm, keepaspectratio=true]{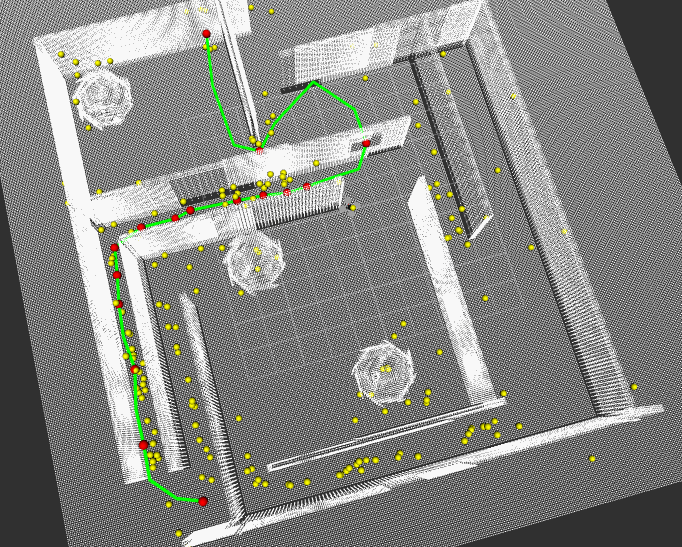}}\hspace{0.25cm}
    \subfloat[$PRM_{RBBS}$\label{Maze_LFBS}]{\includegraphics[height= 4cm,width = 4cm, keepaspectratio=true]{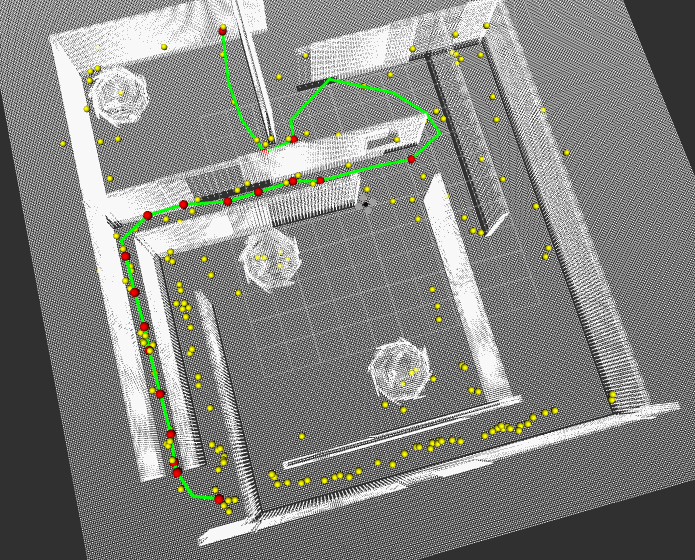}}\\
    \subfloat[]{\includegraphics[width=0.98\columnwidth, height=4cm,keepaspectratio=true]{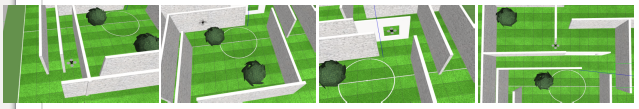}}
    \caption{UAV trajectories when using (a) LFBS  and (b) RBBS. The critical narrow passage samples are shown in yellow. The critical samples within the path graph are shown in red. In (c) snapshots of the UAV motion is shown. The maze environment dimension is $20m\times 20m \times 3m$  }
    \label{fig:maze}
\end{figure}

\begin{figure}
    \centering
    \subfloat[$PRM_{LFBS}$\label{Maze_LFBS}]{\includegraphics[height= 4cm,width = 4cm, keepaspectratio=false]{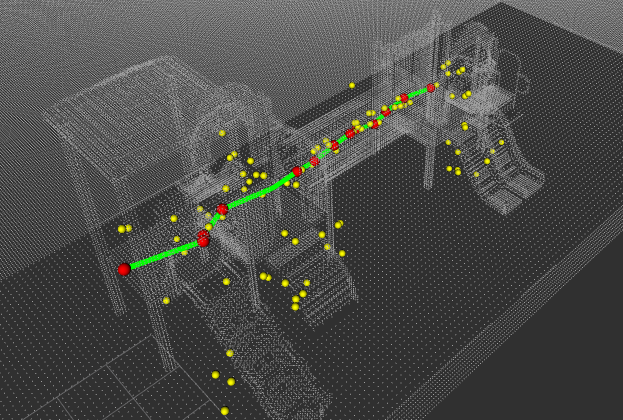}}\hspace{0.25cm}
    \subfloat[$PRM_{RBBS}$\label{Maze_LFBS}]{\includegraphics[height= 4cm,width = 4cm, keepaspectratio=false]{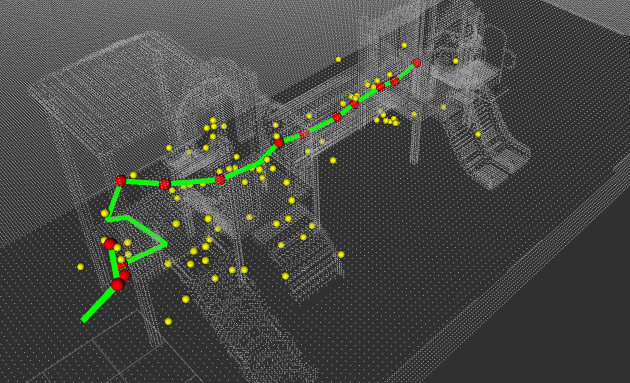}}\\
    \subfloat[]{\includegraphics[width=0.98\columnwidth, height=4cm,keepaspectratio=true]{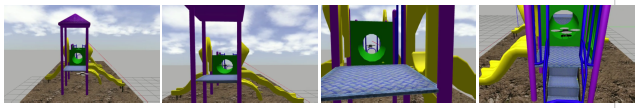}}
    \caption{UAV trajectories when using (a) LFBS  and (b) RBBS. The critical narrow passage samples are shown in yellow. The critical samples within the path graph are shown in red. In (c) snapshots of the UAV motion is shown. The tunnel environment dimension is $20m\times 5m \times 5m$  }
    \label{fig:tunnel}
\end{figure}
\begin{table}
		\centering
		\resizebox{0.48\textwidth}{!}{
		\begin{tabular}{ |p{1.5cm}|p{1.5cm}|p{1.5cm}||p{1.5cm}|p{1.5cm}|  }
        \hline
        \multicolumn{5}{|c|}{LFBS, RBBS samples = 200, prm samples = 300 } \\
        \hline
        Step Size & \multicolumn{2}{|c||}{LFBS} & \multicolumn{2}{|c|}{RBBS}\\
        \hline
         & $t_\varepsilon$(LFBS) & $T_{PRM}$ & $t_\varepsilon$(RBBS) & $T_{PRM}$\\
        \hline
        0.5 &78.49    &298.57    &162.48    &396.83    \\
        \hline
        1 &40.22    &269.91    &102.87    &330.6    \\
        \hline
        \end{tabular}}
\caption{Algorithm execution time comparison results for scenario shown in the Fig.\ref{fig:maze}.}
\label{Table_Maze}
\end{table}
\begin{table} 
		\centering
		\resizebox{0.48\textwidth}{!}{
		\begin{tabular}{ |p{1.5cm}|p{1.5cm}|p{1.5cm}||p{1.5cm}|p{1.5cm}|  }
        \hline
        \multicolumn{5}{|c|}{LFBS, RBBS samples = 100, prm samples = 200 } \\
        \hline
        Step Size & \multicolumn{2}{|c||}{LFBS} & \multicolumn{2}{|c|}{RBBS}\\
        \hline
         & $t_\varepsilon$(LFBS) & $T_{PRM}$ & $t_\varepsilon$(RBBS) & $T_{PRM}$\\
        \hline
        0.2 &74.81    &202.20    &124.56    &253.09    \\
        \hline
        0.5 &46.67    &192.20    &102.6    &248.27    \\
        \hline
        1 &28.40    &172.09    &75.11    &224.65    \\
        \hline
        \end{tabular}}
\caption{Algorithm execution comparison results for scenario shown in the Fig.\ref{fig:tunnel}.}
\label{Table_Tunnel}
\end{table}

\begin{figure}[h]
    \centering
    \subfloat[]{\includegraphics[width=0.98\columnwidth, height=2.5cm,keepaspectratio=true]{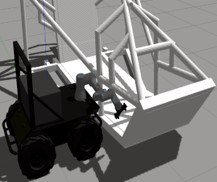}}\hspace{0.25cm}
    \subfloat[]{\includegraphics[width=0.98\columnwidth, height=2.5cm,keepaspectratio=true]{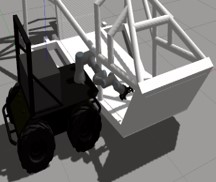}}\\
    \subfloat[]{\includegraphics[width=0.98\columnwidth, height=2.5cm,keepaspectratio=true]{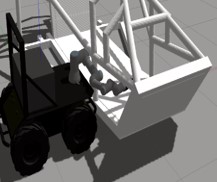}}\hspace{0.25cm}
    \subfloat[]{\includegraphics[width=0.98\columnwidth, height=2.5cm,keepaspectratio=true]{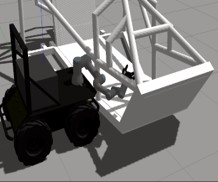}}

\caption{Snapshots of UR3 motion planning for scenarios like welding operation on a chassis with LFBS based PRM }
\label{fig:chasis}
\end{figure}
\begin{table}[h] 
		\centering
		\resizebox{0.35\textwidth}{!}{
		\begin{tabular}{|c|c|c|}
		\hline
		\textbf{Step-size, $a$}& \textbf{Fig.} &\textbf{Time(avg. in s)}\\ 
		\hline
		.5 & Env: Chasis&	233\\
		\hline
		.5 & Env: Window&	253\\
		\hline
		\end{tabular}}
\caption{Time taken in seconds to plan the path for simulated environments as shown in Fig. \ref{Intro_Figure},\ref{fig:chasis}, for a given step size of $0.5$.}
\label{gazebo_table}
\end{table} 

A comparison on algorithmic execution time for both LFBS and RBBS based PRM are shown in Table \ref{Table_Maze}-\ref{Table_Tunnel}. For the Maze scenario, a total of $200$ narrow passage samples are generated for both the RBBS \& LFBS based PRM. These samples are then included within a basic PRM graph network of $300$ nodes. The execution time $t_\varepsilon$ is an average of $300$ trials and an instance of the generated paths is shown in Fig.\ref{fig:maze} for the Maze scenario. Apart from the observation that $t_\varepsilon(LFBS)$ and $T_{PRM}$ for LFBS based PRM are less 
when compared to RBBS based PRM, we also note that for most of the trials the RBBS based PRM could not find a path through the narrow window. One such instances are shown in Fig. \ref{fig:maze}.

Similar results can be seen in case of Tunnel environment. The LFBS based PRM takes lesser computational time with improved sample qualities in the critical constricted region as shown in Fig.\ref{fig:tunnel}. For a large variation in step size the proposed method shows better performance in each cases. We also note that RBBS generates fewer critical samples.

Table \ref{gazebo_table} shows the average time taken to plan the LFBS based PRM trajectories for manipulation motion scenario in constrained environment. Fig.\ref{fig:chasis} shows various snapshots of the executed trajectory. A total number of $200$ LFBS samples are generated for a basic PRM graph of $100$ nodes.
\section{Discussion}\label{discussion}
In this section, we discuss some advantages of the proposed L\'evy flight based bridge sampling technique. It can be noted that, $t_{\varepsilon}$ and $\gamma_c$ for Gaussian Sampler, RWS and LFS are very less, however, the returned sampling quality is quite poor. One possible reason for such can be that these strategies try to sample the configurations in $C_{\text{free}}$, near the boundary of the obstacles, irrespective of there is a narrow region around the obstacle. On the other hand, the RBBS and the proposed LFBS and RWBS, after generating samples around the obstacles, check for the presence of narrow passage using a subsequent bridge test. This additional procedure increases $t_{\varepsilon}$ and $\gamma_c$, but the sampling quality improves substantially. 

%1.) Table for RBBS performance variation with step-size in Bar Shape Obstacle Scenario
\begin{table} 
		\centering
		%\resizebox{0.5\textwidth}{!}{
		\begin{tabular}{|c|c|c|c|c|c|}
		\hline
		\textbf{Step-size, $a$}& \textbf{2}& \textbf{5}& \textbf{10}& \textbf{15}& \textbf{20}\\ 
		\hline
		\textbf{3-Link}&	6.9561&		6.1784&		6.3532&		6.4736&		6.8388	\\
		\hline
		\textbf{5-Link}&	7.4344& 	7.60037&	7.9361&		9.9487&		12.005	\\
		\hline
		\textbf{7-Link}&	10.923&		9.3731&		9.5287&		12.665&		17.128	\\
		\hline
		\end{tabular}%}
\caption{RBBS performance variation, ($t_\varepsilon$, in seconds) with step-size for the Bar Shape Obstacle Scenario. An optimal sample quality for a $5$-link scenario is obtained when $a = 10$, while the same is obtained for $a = 2$ for $7$-link scenario. In both the cases, the co-relation between sample quality and $t_{\varepsilon}$ are inconsistent.}
\label{table1}
\end{table}

One significant advantage offered by the proposed LFBS method is that, unlike RBBS, the corresponding sample quality and algorithmic success probability is independent of the base step size, $a$, of the random walk as evident from the empirical simulations. In case of RBBS, the quality does depend on the value of the bridge length chosen for the bridge test. It gives good results if the length is comparable to or more than the size of the configuration space narrow gap while the sampling quality deteriorates if the length is less than that of the gap and RBBS fails to identify the narrow passage, thus scattering the sample configurations around the obstacle surface. This is also reflected in Table \ref{table1} which shows for various DOF manipulators. Note that as narrow passages in a high dimensional configuration spaces are impossible to construct geometrically, hence the choice of a suitable bridge length ,$a$, has to undergo a trial and error method. Moreover, configurations that are sampled outside the obstacle surface may generate false bridge test positive check (intersection with self) and hence the time/collision-check. Such occurrence can be observed in the results shown in Fig.  \ref{fig:graph1} and \ref{fig:graph3}  Here LFBS and RWBS are able to identify narrow passage successfully with good sampling quality, for all cases irrespective of the step-size at the expense of a moderate increase in execution time/collision checks.

The improved sample quality of LFBS and RWBS, within the key region of a high dimensional configuration space can be attributed to the fact that with the increase in dimension, a random walker explores more than repeats back to the initial state, \cite{bera2010efficient}. This has a direct co-relation with the first passage time and hitting probabilities of the corresponding stochastic process. In case of RBBS, the process is non-Markovian and thus generated samples are clustered about the centroid of a configuration space obstacle. We also observe that, a) in case of comparatively low dimensional \textit{C-space} scenarios and for simple workspace obstacle environments, the $t_{\varepsilon}$, and $\gamma_c$ are less than RBBS b) in higher dimensional \textit{C-space} scenarios and in complex environments where robot maneuverability is more restricted, the sampling quality offered by LFBS or RWBS are better than RBBS, even though time/collision-check might be more. 
\section{Conclusion}\label{conclusion}
In this paper we show L\'evy flight based narrow passage sampling mechanism which shows an improved performance when compared to the state of the art methods. Extensive simulations are performed for a variety of $2$D and $3$D obstacle scenario which establish the advantages over the state of the art. It may be interesting to investigate certain key properties of the stochastic nature of the algorithm, namely the relationship between the success probability of the algorithm and the obstacle's geometry, such as shown in \cite{bera2014analysis}. This we left for future work.

\bibliographystyle{IEEEtran}
\bibliography{References}

\end{document}